\pdfoutput=1
\documentclass[11pt]{article}

\usepackage[review]{EACL2023}

\usepackage{times}
\usepackage{latexsym}

\usepackage[T1]{fontenc}
\usepackage[utf8]{inputenc}

\usepackage{microtype}

\usepackage{inconsolata}

\title{BEVERS: A General, Simple, and Performant Framework \\ for 
Automatic Fact Verification} % should I not use the same title as my thesis? I'm unsure.

\author{Mitchell DeHaven \\
  Information Sciences Institute \\
  University of Southern California \\
  \texttt{mdehaven@isi.edu} \\\And
  Stephen Scott \\
  University of Nebraska-Lincoln \\
  \texttt{sscott2@unl.edu} \\}

\begin{document}
\maketitle
\begin{abstract}
Automatic fact verification has become an increasingly popular topic in recent years and among datasets the Fact Extraction and VERification (FEVER) dataset is one of the most popular. In this work we present BEVERS, a tuned baseline system for the FEVER dataset. Our pipeline uses standard approaches for document retrieval, sentence selection, and final claim classification, however, we spend considerable effort ensuring optimal performance for each component. The results are that BEVERS achieves the highest FEVER score and label accuracy among all systems, published or unpublished. We also apply this pipeline to another fact verification dataset, Scifact, and achieve the highest label accuracy among all systems on that dataset as well. We also make our full code available\footnote{\url{https://github.com/mitchelldehaven/bevers}}.
\end{abstract}

\section{Introduction}

The danger of misinformation online has gained significant attention in recent years. This has been reignited by the recent COVID-19 pandemic, where social media sites and other entities were tasked with identifying misleading content or false content to warn users. Being able to develop systems to automate or build tools to improve this process could reduce the need for human annotators to mark content as being misleading or false. 

The Fact Extraction and VERification (FEVER) dataset \cite{fever-paper} is one the largest and most popular datasets aimed at automated fact verification. The FEVER dataset is comprised of 185,445 claims and uses a 2017 dump of Wikipedia as the corpus to verify the claims, which results in a corpus size of over 5,000,000 articles. For each claim, 
the task is to find the relevant Wikipedia page(s), the relevant sentence(s) within the page(s), and finally given the relevant sentences and claim determine if the claim is supported, refuted, or there is not enough information. As such, a fairly standard pipeline of a document retrieval system, a sentence selection system, and a final claim classification system is used by most of the systems for the task. The primary metric for the dataset is FEVER score. The FEVER score requires both that the predicted label is correct as well as at least one piece of correct evidence being retrieved as predicted evidence.

Much of the recent work has examined parts of the pipeline and made novel improvements over baseline approaches. For our system, rather than making novel improvements against the baseline pipeline, we instead tune each of these components to ensure maximum performance. In fact, our pipeline is quite similar to one of the first FEVER systems to utilize Transformer models \cite{bert-fever}. We call our system Baseline fact Extraction and VERification System (BEVERS). Despite its relative simplicity, our system attains state of the art (SOTA) performance on the FEVER blind test set. When applying our baseline pipeline to another popular fact verification dataset, Scifact \cite{scifact}, our system achieves the highest label F1 score on that dataset as well. 

\section{Related Work and Methods}
\subsection{Document Retrieval}
The initial baseline for FEVER \cite{fever-paper} utilized a standard TF-IDF document retrieval model. \citet{athene} improved on this by using named entity recognition (NER) to extract query terms from the claim text and query those terms against 
WikiMedia's API\footnote{\url{https://www.mediawiki.org/wiki/API:Main\_page}}, which has become widely used among other systems. Recently systems such as those from \citet{stammbach} and \citet{t5-fever} have used a combination of traditional IR approaches with \citeauthor{athene}'s (\citeyear{athene}) NER approach. We follow a similar setup, however, we replace the approach of 
\citeauthor{athene}'s (\citeyear{athene})
%\citet{athene} 
use of WikiMedia's API. We similarly extract named entities to form query terms, however, we run those against a fuzzy string search system using the titles of the documents. 
For our TF-IDF, we build separate representations for documents and titles. This is for two reasons. First, it allows us to separately optimize the parameters for titles and documents. 
Second, it forces the retrieval system to give titles more attention as it is forced to retrieve half of all documents based on the title alone. We give an ablation over these design decisions in Appendix \ref{sec:ablation_studies}.

\subsection{Sentence Selection}
After retrieving documents, the next step is to score evidence and form a ranking for the predicted evidence of the claim. The simplest approach to doing this is referred to as ``point-wise'' ranking, in which each sentence is scored individually against the claim. This is the approach utilized by most systems. \citet{bert-fever} looked at improving on this utilizing a pairwise approach to ranking. \citet{stammbach} found that utilizing document-wide context via sparse attention Transformers improves on point-wise approaches. Our system utilizes a simple point-wise approach to sentence selection to form the predicted evidence. We look at two cases, treating the task as both a binary classification task and a ternary classification task. In the binary case, the label set is simply RELEVANT and IRRELEVANT with the softmax score of RELEVANT being used for ranking. In the ternary case, we use REFUTES, NOT ENOUGH INFO, and SUPPORTS as the labels and use $1 -$ NOT ENOUGH INFO softmax score for ranking. We randomly sample sentences from the document retrieved from our document retrieval approach for negative samples. In the binary case, these random negative samples are assigned to the IRRELEVANT label class and all true evidence is assigned to RELEVANT. In the ternary case, the negative samples are assigned to NOT ENOUGH INFO, and the true evidence is assigned to its respective labels, REFUTES and SUPPORTS.

In addition, we utilize a process we call evidence-based re-retrieval. The FEVER dataset includes hyperlink information for each sentence in the dataset. This process takes the initial set of predicted evidence for a claim and extracts additional documents based on hyperlinks found in the initial sentences retrieved. Sentences from these additional documents are scored and combined with the initial sentences to form a final top 5 predicted evidence. This process is very similar to \citeposs{stammbach} ``multi-hop retrieval'', with slight differences in how sentences are discounted when combining the two sets of sentences. Stammbach sets evidence from re-retrieved documents just above a predefined threshold for selection to prevent re-retrieved evidence from pushing evidence from the initial retrieval outside of the top 5. We similarly found that simply combining both sets together actually hurts recall, because evidence from re-retrieval sometimes pushes out relevant evidence from the initial retrieval. In our approach, we scale the sentence selection scores of the re-retrieved sentences by the score of the original sentence that was responsible for its retrieval. Thus, if evidence $s_j$ was retrieved due to a hyperlink in $s_i$ the final retrieval score is $score(s_i) \times score(s_j)$. Scaling this way reduces re-retrieved evidence pushing evidence from the initial retrieval from the top-5 selection. It also allows re-retrieved evidence scores to be proportional to the score of the initial evidence responsible for its retrieval. 

\subsection{Claim Classification}
The claim classification portion has recently seen the most diversity in approaches to the task. The initial Transformer approach of \citet{bert-fever} formed predictions for each claim and evidence pair, using a simple set of rules to aggregate labels across the different pieces of evidence. Subsequently,  several works examined the use of graph neural networks as the claim classification model \cite{kgat, dream}, showing improvements over simply using Transformers due to their ability to aggregate information over different pieces of evidence. More recently, increasing the size of the Transformer models and concatenating all evidence sentences together have shown further improvements, with \citet{t5-fever} using T5 \cite{t5} and \citet{stammbach} using DeBERTa V2 XL MNLI \cite{deberta}. Finally, the previous SOTA among public systems, ProofVER \cite{proofver}, utilizes natural language proofs generated via seq2seq models for interpretable inference steps.

For our approach, we look at prediction over singleton, concatenated, and a mixed case. We predict a top-5 evidence set for each claim for training using our document selection and sentence selection. In the singleton case, we generate a prediction for each piece of evidence using as input the $\langle$claim, evidence$\rangle$ pairs. In the concatenated case, we concatenate all the evidence together and form the input based on $\langle$claim, evidence$_1$, evidence$_2,  \ldots \rangle$. For the mixed approach, we mix the singleton approach and concatenated approach together. For the singleton and mixed approach, we have multiple predictions for each claim. To aggregate these into a single score, we use the softmax scores for each prediction with the retrieval scores and train a gradient boosting classifier \cite{gradient_boosting} on these inputs to produce a single prediction. In the singleton case, the input is a $5 \times 4$ matrix (5 pieces of evidence, 3 softmax scores and a retrieval score). In the mixed case, the input is a $6 \times 4$ matrix (includes the additional concatenated input softmax scores and the retrieval score, computed from the average retrieval scores of the 5 pieces of evidence). The singleton and concatenated approaches have been used previously \cite{bert-fever, t5-fever}, while we are not aware of any works that look at simply mixing these approaches together. 

\section{Experimental Setup}
\begin{table}
\centering
\begin{tabular}{lc}
\hline
\textbf{Hyperparameter} & \textbf{Values}\\
\hline
Force Lowercase & True, False \\
Force ASCII &  True, False \\
Norm &  L2, None\\ 
Sublinear TF &  True, False \\ 
Max Ngram &  1, 2 \\\hline
\end{tabular}
\caption{The hyperparameter search for our TF-IDF system.}
\label{tab:tf_idf_param}
\end{table}

\begin{table}
\centering
\begin{tabular}{lc}
\hline
\textbf{Hyperparameter} & \textbf{Values}\\
\hline
Label Set & binary, ternary \\
Negative Samples &  5, 10, 20, 40 \\\hline
Learning Rate & 1e-5, 6e-6, 3e-6 \\
Label Smoothing &  0.0, 0.1, 0.2 \\\hline
\end{tabular}
\caption{The hyperparameter search our sentence selection model.}
\label{tab:sentence_selection_params}
\end{table}

What we believe to be the source of improvements for our system is hyperparameter tuning for each component. We identify hyperparameters and potential values and run a grid search to find the optimal configurations for each component. In this section, we will go over each of the grid searches providing additional details on the exact setup. 

For our TF-IDF system, we utilize SciKit Learn's \cite{scikit-learn} TF-IDF representation. In Table~\ref{tab:tf_idf_param} we list the hyperparameters and their candidate values used in the grid search. We use recall @ 5 on the development set for finding the best configuration. The fuzzy string search is implemented using Sqlite's spellfix1 virtual table\footnote{\url{https://www.sqlite.org/spellfix1.html}}. We set a simple edit distance threshold for retrieving additional documents.

Our sentence selection hyperparameter tuning is split into two sections. First, we optimize the number of negative samples selected as well as binary vs ternary classes for ranking. Since the FEVER dataset does not provide evidence for NOT ENOUGH INFO claims, negative samples must be used to generate training examples for these. Using the best selection from the initial setup, we tune the learning rate and label smoothing. The candidate values for the tuning can be found in Table~\ref{tab:sentence_selection_params}.  Given the imbalance in the training set and the balanced nature of the dev and test set, we oversample the minority classes so that label distribution in the training set matches that of the dev and test sets. We use the dev set for determining optimal hyperparameter values. RoBERTa Large \cite{roberta} is used as the initial model for fine-tuning. 

The claim classification tuning setup is quite similar to sentence selection. We initially tune the learning rate and label smoothing using the same candidate values for the concatenated case. Instead of tuning the model types of singleton, concatenated, and mixed, we simply use the best hyperparameter configuration and train a model for each of these to draw final comparisons. Again, given the imbalance in classes in the train set, we use class weighting to compensate for this imbalance. For fine-tuning we use RoBERTa Large MNLI and DeBERTa V2 XL MNLI.

\begin{table}
\centering
\begin{tabular}{lc}
\hline
\textbf{Hyperparameter} & \textbf{Values}\\
\hline
Learning Rate & 0.1, 0.3 \\
Estimators & 20, 40, 60, 80, 100\\
Max Depth &  2, 4, 6, 8 \\\hline
\end{tabular}
\caption{The hyperparameter search our gradient boosting model.}
\label{tab:xgbc_params}
\end{table}

\begin{table*}
\centering
\begin{tabular}{lcc}
\hline
\textbf{System} & \textbf{Test LA} & \textbf{Test FEVER}\\
\hline
\citet{bert-fever} & 71.86\% & 69.66\% \\
KGAT \citet{kgat} &  74.07\% & 70.38\% \\
LisT5 \citet{t5-fever} & 79.35\% & 75.87\% \\
\citet{stammbach} & 79.20\% & 76.80\% \\
ProoFVer \citet{proofver} & 79.47 \% & 76.82\% \\
Ours (RoBERTa Large MNLI) singleton & 78.01\% & 76.09\% \\
Ours (RoBERTa Large MNLI) concatenated & 79.14\% & 76.69\% \\
Ours (RoBERTa Large MNLI) mixed & 79.39\% & 76.89\% \\
Ours (DeBERTa V2 XL MNLI) mixed & \textbf{80.24}\% & \textbf{77.70}\% \\\hline
\end{tabular}
\caption{Full system comparison for label accuracy (LA) and FEVER score on the blind test set.}
\label{tab:system_results}
\end{table*}

Finally, for the singleton and mixed approaches, we use XGBoost \cite{xgboost} for training a classifier to aggregate the predictions into a single prediction. Similarly, we define a hyperparameter grid to find the optimal values. Since the previous steps were all trained on the train set and thus the softmax scores and retrieval scores will be overly optimistic on the training set, we instead train the XGBoost classifier on the dev set. We use 4-fold cross-validation to find the optimal configuration. 

\section{Results}

 \begin{table}[!h]
\centering
\begin{tabular}{lc}
\hline
\textbf{System} & \textbf{Dev Recall @ 5}\\
\hline
\citet{athene} & 87.10\% \\
\citet{kgat} &  94.37\% \\%\footnote{Although we marginally outperform KGAT on the dev set, our system achieves 93.42\% recall @ 5 on the test set against their 87.47\%. Unfortunately many systems do not report recall @ 5 on the test set, so our comparison must use dev.} \\
\citet{bert-fever} & 88.38\% \\
\citet{t5-fever} & 90.54\% \\
\citet{stammbach} & 93.62\% \\
Ours & 92.03\% \\
   + re-retrieval & \textbf{94.41}\% \\\hline
\end{tabular}
\caption{The results of several sentence selection systems in terms of recall @ 5 on the dev set.}
\label{tab:sentence_selection_results}
\end{table}

For sentence selection, the primary metric used is recall @ 5. This is due to the fact that when computing FEVER score, the scoring metric will only consider up to 5 pieces of predicted evidence. 
In Table~\ref{tab:sentence_selection_results} we compare our sentence selection system against several other top systems on the dev set. As can be seen, our sentence selection system outperforms all previous systems in terms of recall @ 5 on the dev set. This is despite using a substantially smaller model relative to \citeposs{t5-fever} T5 approach as well as only using pointwise scoring for sentence selection as opposed to \citeposs{stammbach} full document context approach. We separate our results from using initial retrieval and including evidence-based re-retrieval, which shows a very large improvement in recall by doing re-retrieval, consistent with \citeposs{stammbach} findings.

For claim classification results, we present the entire end-to-end results for our system in Table~\ref{tab:system_results}. The simple approach of mixing the singleton and concatenate approaches gives a small improvement, although is not a substantial source of improvement. Despite the singleton approach being incapable of modeling claims that require multi-hop evidence, it still performs well.
Despite using a relatively smaller model of 300 million parameters when compared to 3 billion with T5 and 900 million with DeBERTa V2 XL MNLI, our RoBERT Large MNLI system achieves the highest FEVER score among all published systems. When we utilize DeBERTa V2 XL MNLI using our mixed approach, we achieve the highest label accuracy and FEVER score amongst all systems, published or unpublished, on the blind test set.

\section{Beyond FEVER: Scifact}

\begin{table}[h]
\centering
\begin{tabular}{lcc}
\hline
\textbf{System} & \textbf{SS + L} & \textbf{Abstract LO} \\
\hline
\citet{VerT5erini} & 58.8 & 64.9 \\
\citet{arsjoint} & 63.1 & 68.1 \\
\citet{multivers} & \textbf{67.2} & 72.5 \\
Ours & 58.1 & \textbf{73.2} \\\hline
\end{tabular}
\caption{System comparison for SS + L F1 score and Abstract LO F1 score on SciFact blind test set.}
\label{tab:scifact_results}
\end{table}

To test this pipeline for automatic fact verification beyond the FEVER dataset, we also apply these methods to the SciFact dataset \cite{scifact}. SciFact is very similar in structure to the FEVER dataset, however, the corpus is composed of scientific articles. A source of difficulty is that claims are often phrased in lay terms, which can be a stark difference in form from how topics are presented in scientific articles. 
The overall size of the dataset is quite a bit smaller as well, containing only 1,409 claims and 5,183 article abstracts, which serve as the corpus. Despite this, we keep our pipeline nearly identical to FEVER, excluding only the fuzzy string search component. We follow the approach of \citet{multivers} for improving the initial models for finetuning given the low resource nature of the dataset.

We show the results of our pipeline in Table~\ref{tab:scifact_results}
compared to the current SOTA \cite{multivers} and other top systems. The metrics reported are sentence selection + label (SS + L) and abstract label only (Abstract LO). These metrics roughly correspond to FEVER Score and label accuracy for FEVER. As can be seen in the SS + L metric, the simplicity of our document retrieval system appears to hold the overall system back. Our system only uses TF-IDF whereas the three others add neural re-rankers on top of their retrieval. Despite this, on the Abstract LO metric our system achieves the highest F1 score on the blind test set, outperforming the SOTA on this metric. 

\section{Conclusion}

We presented BEVERS, a strong baseline approach for the FEVER and SciFact datasets. Despite being similar to previous works in structure \cite{bert-fever} and utilizing little in terms of novel improvements, our system was able to achieve SOTA performance on FEVER and the highest label accuracy on SciFact. We primarily attribute these improvements to diligent hyperparameter tuning and error analysis. While several previous works have shown novel contributions to portions of the pipeline can yield improvements, in this work we show a well-tuned baseline is very strong.

\section{Limitations}

As shown with SciFact, this pipeline struggles in situations where there is a mismatch in how claims are phrased and how evidence is phrased in the corpus. Since our retrieval method is term-based, synonymous terms are often missed, and thus in such systems utilizing neural retrieval methods will offer better performance. In addition, this work does not thoroughly examine which design decisions or approaches led to the improvements seen in this pipeline. We note that evidence-based re-retrieval does give substantial improvements, yet even without re-retrieval, our sentence selection outperforms most previous systems by a substantial margin, so it is not the sole source of improvement.
\bibliography{anthology,custom}
\bibliographystyle{acl_natbib}

%\appendix

%\section{Example Appendix}
%\label{sec:appendix}
%
%This is a section in the appendix.
\appendix

\section{Optimal Hyperparameter Settings}
\label{sec:optimal_hyperparameters}

In Table \ref{tab:optimal_hyperparamters_tfidf} we show the optimal hyperparameter settings for the various TF-IDF configurations. To minimize space, we use "Cat" to refer to the concatenated TF-IDF setup. In Table \ref{tab:optimal_hyperparamters_ss} and Table \ref{tab:optimal_hyperparamters_cc} we show the optimal hyperparameter values for sentence selection and claim classification models. Finally, in Table \ref{tab:optimal_hyperparamters_xgboost} we include the optimal hyperparameter values for the XGBoost classifier. 
\begin{table}[!h]
\centering
\begin{tabular}{lcc}
\hline
\textbf{Hyperparameter} & \textbf{Cat} & \textbf{Title}, \textbf{Document} \\
\hline
Force Lowercase & True & True, False \\ 
Force ASCII & True &  True, True \\
Norm & None & L2, None \\
Sublinear TF & True & True, True \\
Max Ngram & 2 & 2, 2 \\
\hline
\end{tabular}
\caption{Optimal hyperparameters for the concatenated and separated TF-IDF configurations.}
\label{tab:optimal_hyperparamters_tfidf}
\end{table}

\begin{table}[!h]
\centering
\begin{tabular}{lc}
\hline
\textbf{Hyperparameter} & \textbf{Optimal Value} \\
\hline
Label Set & Ternary \\ 
Negative Samples & 10 \\
Learning Rate & 3e-6 \\
Label Smoothing & 0.0 \\
\hline
\end{tabular}
\caption{Optimal hyperparameters for sentence selection model.}
\label{tab:optimal_hyperparamters_ss}
\end{table}

\begin{table}[!h]
\centering
\begin{tabular}{lc}
\hline
\textbf{Hyperparameter} & \textbf{Optimal Value} \\
\hline
Learning Rate & 3e-6 \\
Label Smoothing & 0.2 \\
\hline
\end{tabular}
\caption{Optimal hyperparameters for claim classification model.}
\label{tab:optimal_hyperparamters_cc}
\end{table}

\begin{table}[!h]
\centering
\begin{tabular}{lc}
\hline
\textbf{Hyperparameter} & \textbf{Optimal Value} \\
\hline
Max Depth & 2 \\
Number of Estimators & 60 \\
Learning Rate & 0.3 \\
\hline
\end{tabular}
\caption{Optimal hyperparameters for XGBoost aggregation classifier.}
\label{tab:optimal_hyperparamters_xgboost}
\end{table}

\section{Ablation Studies}
\label{sec:ablation_studies}

In Table \ref{tab:document_retrieval_ablation} we show the impacts of various design choices for document retrieval and their impacts on sentence selection. We use our best sentence selection model for ranking the sentences retrieved by the document retrieval approaches. Previous works use OFEVER from the original paper as a metric for comparing document retrieval methods, however, OFEVER does not account for different approaches retrieving different numbers of documents given that is an oracle approach. Thus, we find measuring the sentence selection in this way gives a better representation of improvements. 

\begin{table}[!h]
\centering
\begin{tabular}{lc}
\hline
\textbf{Retrieval Approach} & \textbf{Dev Recall $@$ 5}\\
\hline
TF-IDF (concatenated) &  84.49 \% \\
  + fuzzy string search & 91.35 \%\\
  + document re-retrieval & 93.58 \% \\ 
TF-IDF (separated) & 87.09 \% \\
  + fuzzy string search & 92.03 \%\\
  + document re-retrieval & \textbf{94.41}\% \\ 
\hline
\end{tabular}
\caption{Dev set recall $@$ 5 using various document retrieval approaches.}
\label{tab:document_retrieval_ablation}
\end{table}

\begin{table*}[!h]
\centering
\begin{tabular}{lcc}
\hline
Author (Model) & Test LA & Test FEVER \\ \hline
KGAT (RoBERTa Large) \cite{kgat} & 74.07 \% & 70.38 \% \\ 
KGAT (CorefRoBERTa) \cite{kgat_coref} & 75.96 \% & 72.30 \% \\ 
Ours (RoBERTa Large) & 76.60 \% & 73.21 \% \\
Ours (RoBERTa Large MNLI) & 77.95 \% & 74.08 \% \\ 
\hline
\end{tabular}
\caption{Comparison between KGAT's claim classification and ours. We use KGAT's released outputs for evidence retrieval, so differences in performance are not attributable to improvements in our system's retrieval approach. }
\label{tab:kgat_comparsion}
\end{table*}

In Table \ref{tab:kgat_comparsion} we compare our claim classification setup with KGAT's. Rather than utilizing our document retrieval and sentence selection, we use KGAT's sentence selection outputs which they make publicly available. This allows for a more direct comparison since we are using the same evidence for forming predictions. The only changes we make: re-score the top 5 evidence from KGAT's sentence selection using our own best sentence selection model and re-train the gradient boosting classifier. Despite using the same evidence as KGAT, our claim classification still outperforms using either RoBERTa Large or RoBERTa Large MNLI. So while some of the improvement in our system is attributable to improvements in document retrieval and sentence selection our approach to claim classification still outperforms previous systems when using the same retrieval outputs.

\end{document}